\documentclass[letterpaper]{article} 
\usepackage{aaai2027}  
\usepackage[hyphens]{url}  
\usepackage{graphicx} 
\usepackage{natbib}  
\usepackage{caption} 
\usepackage{booktabs}
\usepackage{threeparttable}
\usepackage{multirow}
\usepackage{array}
\usepackage{amsmath}
\usepackage{amssymb}

\def\E{\textsc{Emergent}}
\def\U{\textsc{Urgent}}
\def\R{\textsc{Routine}}

\newcommand{\kap}{\ensuremath{\kappa}}

\newcommand{\code}[1]{\texttt{#1}}

\newcommand{\bigqwen}{Qwen3-235B}

\newcommand{\Phirule}{\ensuremath{\Phi}}
\newcommand{\Phihat}{\ensuremath{\hat{\Phi}}}

\newcommand{\cref}[1]{Section~\ref{#1}}
\newcommand{\Cref}[1]{Section~\ref{#1}}

\newif\ifclinicaldata
\clinicaldatafalse

\title{Guideline-as-Oracle: Zero-Annotation Training of an Ophthalmic Telephone Triage Agent}

\author{
    Chenyu Wang\textsuperscript{\rm 1,3}\equalcontrib,
    Yi Liu\textsuperscript{\rm 2}\equalcontrib,
    Baoqing Li\textsuperscript{\rm 1},
    Min Tu\textsuperscript{\rm 1},
    Diping Song\textsuperscript{\rm 3}\corresponding
}

\affiliations{
    \textsuperscript{\rm 1}Shanghai Institute of Microsystem and Information Technology,
    Chinese Academy of Sciences\\
    \textsuperscript{\rm 2}Independent Researcher\\
    \textsuperscript{\rm 3}Shanghai Artificial Intelligence Laboratory
}

\begin{document}
\maketitle

\begin{abstract}
Scaling supervision for multi-turn medical agents is difficult because expert dialogue annotation is costly and clinical conversations are privacy-restricted. We introduce \emph{Guideline-as-Oracle} (GAO), which compiles American Academy of Ophthalmology guidance into a 70-row operational rule table and uses it as the sole source of instance-level supervision for 3,000 training dialogues, reserving human labeling for evaluation. Because converting rules into dialogues is itself a design problem, we catalog eight construction strategies—including cited-row tier assignment, one-fact boundary pairs, metadata-only repair, and label repair—and characterize the evidential status of each: labeling mechanism, null, confounded, or evaluated only as a package. Fine-tuning a 9B backbone on this corpus yields GAO-Triage, improving agreement with a 201-case operational reference from 61.7\% to 74.1\% (exact McNemar $p{=}0.0046$) and emergent-case recall from 9.5\% to 69.0\%; the gains persist across a second seed and patient simulator. None of the seven general-purpose systems we test dominates GAO-Triage on both metrics, and GAO-Triage requires no frontier model at inference time. Permuting label–dialogue assignments collapses the model to a constant-routine predictor, indicating that the signal lies in guideline-derived assignment rather than dialogue surface form. Label repair coincides with the disappearance of a late-training safety degradation. 
\end{abstract}


\section{Introduction}
\label{sec:intro}

Ophthalmic telephone triage must determine from verbal interaction alone,
without examination or imaging, whether a caller requires emergent (E), urgent
(U), or routine (R) care. Under-triage can delay management of
sight-threatening conditions; over-triage consumes clinical capacity. An
automated triage agent must therefore acquire discriminating evidence over
multiple turns and emit an auditable disposition, rather than classify a static
vignette.

\label{sec:motivation}

Expert annotation of multi-turn call transcripts does not scale, and
identifiable clinical conversations are rarely releasable as training data.
Specialty societies, however, publish structured guidance that maps
presentations to disposition tiers. This leaves two questions: what does it take
to compile such a rule table into supervision for a competent multi-turn agent,
and what in that supervision carries the resulting capability? We study both with
an author-compiled operationalization as the sole instance-level supervision for
training dialogues, reserving human labeling for evaluation.

\begin{figure}[t]
\centering
\includegraphics[width=\columnwidth]{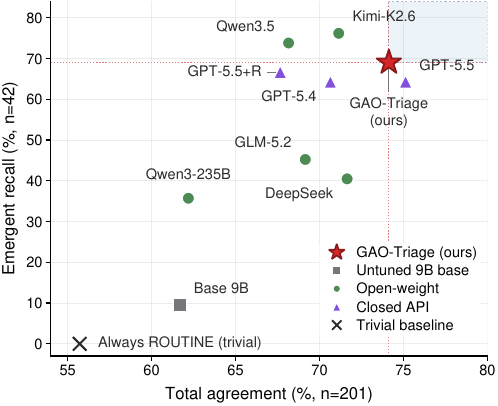}
\caption{Clean-protocol results ($n{=}201$; \emph{clean} means the
complete-profile heuristic of \cref{sec:eval} is disabled): operational-reference
agreement against recall on 42 emergent cases. Dotted crosshairs mark GAO-Triage
(red star), and the shaded upper-right quadrant holds any arm that would beat it
on \emph{both} axes. None does. Base and ROUTINE-only anchor the plot;
GPT-5.5$+$R is GPT-5.5 given the 70-row table. Remaining checks appear in
Table~\ref{tab:headline} and the supplement. Nonsignificance is not
equivalence.}
\label{fig:results}
\end{figure}

\begin{figure*}[t]
\centering
\includegraphics[width=\textwidth]{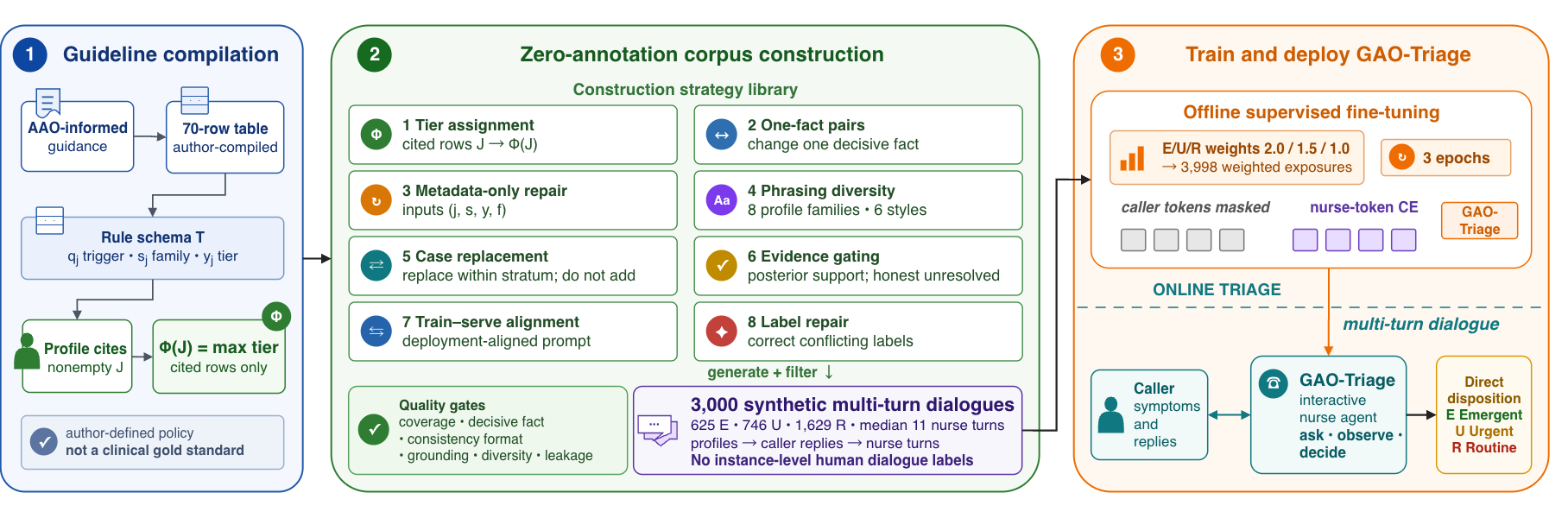}
\caption{Guideline-as-Oracle (\cref{sec:problem}). \textbf{(1)} Authors compile
AAO-informed guidance into a 70-row table; a profile cites a nonempty row set
$J$ and receives $\Phirule(J)$, the maximum tier over the cited rows only.
\textbf{(2)} Eight construction strategies turn that table into 3{,}000
multi-turn dialogues carrying no instance-level human labels; the quality gates
are an intended taxonomy, not a verified property. \textbf{(3)} Class-weighted
resampling and nurse-token cross-entropy yield GAO-Triage, which then runs as an
interactive nurse and states a disposition directly. Human labels enter only at
evaluation. Because the compiled policy is author-defined rather than a clinical
gold standard, downstream scores measure policy conformance.}
\label{fig:system}
\end{figure*}

We compile American Academy of Ophthalmology (AAO) materials into a 70-row
operational table and let it supply every instance-level dialogue label
(\emph{Guideline-as-Oracle}, GAO), writing GAO-Triage for the 9B backbone
fine-tuned on the resulting corpus. Both names are narrower than they sound:
\emph{zero annotation} excludes training labels, not the compilation and audit
of the policy itself, and \emph{Oracle} is a construction role, not a clinical
gold standard or an AAO endorsement.

We evaluate full-parameter supervised fine-tuning (SFT) of that backbone on a
201-case operational reference, which also informed rule coverage and model
selection---so the scores below measure conformance to a policy under
development, not held-out clinical generalization (\cref{sec:eval}). Auditing
our own evaluation pipeline surfaced benchmark-derived paraphrases among
repaired dialogues and a complete-profile heuristic that could overwrite a
verdict from hidden case data; every primary claim rests on clean comparisons
with that heuristic disabled (\cref{sec:clean-rerun}).

This paper makes three contributions:

\begin{itemize}
    \item \textbf{A design space for rule-to-dialogue construction.}
    We systematize eight strategies for converting a rule table into training dialogues: cited-row labels, one-fact boundary pairs, metadata-only repair, phrasing diversity, replacement, evidence gating, train--serve alignment, and label repair. We further characterize the evidential status of each strategy. Although their individual effects are not separately identified, the resulting null findings and confounds delineate what can---and cannot---be concluded from the available evidence (\cref{sec:strategies}).

    \item \textbf{A nurse-role agent competitive with frontier models.}
    SFT increases agreement from 61.7\% to 74.1\% (exact McNemar $p{=}0.0046$) and emergent recall from 9.5\% to 69.0\%. Across eight external nurse arms, none dominates GAO-Triage on both agreement and emergent recall (\cref{sec:clean-rerun}).

    \item \textbf{A controlled localization of the learned capability.}
    Permuting labels across dialogues while preserving the dialogue texts collapses the model to a constant \emph{routine} prediction ($112/201$), with emergent recall falling to $0/42$. This intervention shows that the learned capability depends on the assignment between dialogues and labels, rather than on dialogue surface form alone (\cref{sec:analysis}).
\end{itemize}

\section{Related Work}
\label{sec:related}

\paragraph{Supervision without per-instance labels.}
Data programming replaces per-instance annotation with user-authored labeling
functions, aggregated by a generative label model
\citep{ratner2016data,ratner2017snorkel}, while a parallel line derives
supervision from written principles or model outputs
\citep{bai2022constitutional,wang2023selfinstruct,taori2023alpaca} and recent
medical work synthesizes agent trajectories from structured domain knowledge
\citep{medresearcher2025}. Our setting differs in what plays the role of the
labeling function. A specialty guideline is not a noisy heuristic to be
denoised: it is the operative decision rule, so a single compiled table can
carry every instance-level label without a label model or majority vote over
weak sources. What we give up is the correctness guarantee a validated clinical
rule would provide, and because our labels are generated rather than aggregated,
the failure mode differs---instead of noisy-but-independent votes, we face
internally contradictory rules, which is why label repair appears as a
construction strategy in its own right (\cref{sec:strategies}). Compiling from
guidance is attractive here precisely because the data that make multi-turn
triage hard to supervise are the data that cannot be released: openly available
clinical resources are records, waveforms, or summaries rather than
conversations
\citep{goldberger2000physionet,johnson2023mimiced,zhao2023pmcpatients}, and
telephone triage calls are rarely shareable even de-identified.

\paragraph{Learning to ask, and learning from outcomes.}
When interaction data are unavailable, task-oriented dialogue systems train
against simulated users \citep{schatzmann2007agenda,kreyssig2018neural}, and
recent LLM work trains proactive questioning under missing information
\citep{huang2025proactive,li2025alfa}. Our boundary-pair strategy pursues the
same goal by a different route: instead of rewarding information gain, it
derives local contrasts from an operational rule, so that two dialogues
differing in one decision-critical fact carry different tiers. Removing these
pairs is null on our reference (\cref{sec:ablations}), so we present it with a
stated null rather than a demonstrated mechanism. Rewarding the outcome directly
is the other obvious route: preference optimization and policy-gradient methods
align models to outcome or preference signals
\citep{ouyang2022instructgpt,rafailov2023dpo,shao2024deepseekmath}, but triage
exposes a credit-assignment gap for them, since the rewarded quantity is the
final disposition whereas the behavior that produces it is what gets asked and
when to stop. Our post-training arm is consistent with that gap---rewarding the
outcome degrades the behavior that earns it 

\paragraph{Interactive clinical agents and their evaluation.}
Medical LLM evaluation has emphasized diagnostic QA and static vignettes
\citep{nori2023gpt4,singhal2023medpalm,thirunavukarasu2023llm}, with self-play
used to elicit richer behavior \citep{tu2024amie}; triage benchmarks, including
ophthalmic ones, remain full-information
\citep{edtriage2025,mittal2026ophthalmic}. Agents that withhold findings until
queried move closer to the telephone setting
\citep{schmidgall2024agentclinic,nori2025sequential}, and one deployed telephone
agent restricts its LLM to scripted preparation under per-call nurse review
\citep{kini2026sofiya}. None targets ophthalmic triage, and none audits its own
evaluation pipeline for the paths we found necessary to check---which is where
the rest of this paper's caution comes from. Synthetic-data pipelines can
transmit benchmark content into training through repair and regeneration
\citep{contamination_survey2024,zhou2023dontmake}, which is why our repair
strategy is specified on metadata alone; where a destructive recorder overwrote
direct verdicts, reconstruction is not identified from retained metadata, so we
bound the affected quantity instead, following work on coarsened and
contaminated data
\citep{heitjan1991coarse,horowitz1995contaminated,manski2003partial}. Our
labeling workflow also uses rule-conditioned LLM audits, whose known biases as
judges \citep{zheng2023judging} are the reason they propose rather than assign
labels. 

\section{Method: Guideline-as-Oracle}
\label{sec:problem}

We model guideline-anchored triage as sequential evidence acquisition under an
author-defined construction policy, not a validated clinical rule.


\subsection{Compiling the Guideline}
\label{sec:compile}

The compiled artifact is an ordered table of rules
$\mathcal{T}=\{(q_j,s_j,y_j)\}_{j=1}^{m}$, each pairing a textual trigger $q_j$
and symptom family $s_j$ with the tier $y_j$ it warrants. We expanded an initial
64-row AAO-informed table \citep{aao2025ppp} to $m=70$ after benchmark analysis,
so coverage is benchmark-informed though no case text was copied; the table
carries neither independent clinician validation nor a mapping to source
passages.

Labeling is by citation rather than by matching. Each generated profile records
the nonempty set $J$ of rows its generator invoked and inherits the most severe
tier among them, $\Phirule(J)$, so a label is only ever as good as the rows the
generator names. The same handle drives two further strategies: a one-fact
boundary pair edits a designated fact together
with its target row, which moves $\Phirule$ by construction, and repair
regenerates a dialogue from its rule, family, tier, and failure tag alone,
withholding the rejected text, the evaluation narrative, and the case identity.
Validators enforce row membership, the maximum-cited-tier rule, and limited
explicit consistency, but not semantic coverage of every trigger, and audits
later found historical violations of the repair specification. Admissible dialogues close at $\Phirule(J)$, and the
\code{Reason} field the nurse emits alongside its tier may cite only findings the
caller supplied; training weights the classes by deterministic resampling and
scores cross-entropy on nurse tokens only.
The diagnostics of \cref{sec:eval-decomp} need one further, deliberately weaker
object. A 39-entry subset over a 25-key schema forms an \emph{executable} matcher
$\Phihat(x)=r(g(x))$: $g$ extracts whichever schema fields a case narrative $x$
mentions, and $r$ runs the corresponding rule subset over them. Being incomplete,
$\Phihat$ never labels a training row or supplies an evaluation
label; it exists so that a case can be read three ways---the disposition
$v(h)$ a system states from the transcript $h$ it elicited, and $\Phihat$ on that
transcript versus on the complete narrative.

\subsection{From Table to Corpus}
\label{sec:corpus}
\label{sec:data}

Because $\Phirule$ supplies every disposition, the corpus carries no
instance-level dialogue labels (Figure~\ref{fig:system}). It was not built in one
pass: class ratio, quality gates, coverage, provider mix, and upsampling all
moved adaptively across successive corpora, so what follows is a development
trace that on its own identifies no intervention effect.

\paragraph{Generation pipeline.}
Profiles span eight symptom families, some of them abstracted from public Reddit
narratives, and dialogue is generated in a separate pass. Distinct roles produce the profile,
the patient replies, and the nurse turns, with family-specific system blocks
supervising every nurse turn. Beyond the validators of \cref{sec:compile},
seven gates screen the output for coverage, decisive-fact support, consistency,
format, grounding, diversity, and leakage; they evolved across batches and
describe an intended taxonomy rather than a verified property.

\begin{figure*}[t]
\centering
\includegraphics[width=0.98\textwidth]{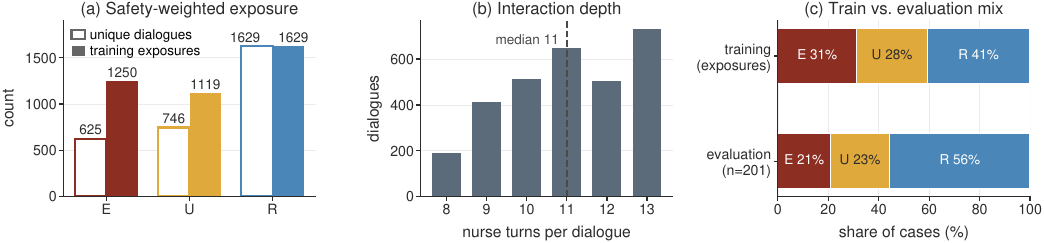}
\caption{Primary boundary-revised corpus. \textbf{(a)} Unique dialogues and
safety-weighted training exposures. \textbf{(b)} Nurse-turn depth (median 11).
\textbf{(c)} Class mix of the training exposures against that of the evaluation
reference, which differ by design: resampling upweights the severe tiers. E/U/R
denote emergent/urgent/routine.}
\label{fig:corpus}
\end{figure*}

For the primary base--SFT comparison, the corpus contains 3{,}000 dialogues
with a median of 11 nurse turns (625 emergent, 746 urgent, and 1{,}629 routine)
after within-stratum replacement of 160 urgent-boundary dialogues. Resampling
with class weights 2.0, 1.5, and 1.0 yields 3{,}998 exposures (Figure~\ref{fig:corpus});
training runs for three epochs with nurse-token supervision and caller-token
masking (\cref{sec:clean-rerun}). Two corpora sit outside this main line: the
optional Group Relative Policy Optimization (GRPO) stage adds no labels but its
starting SFT weights were not retained (\cref{sec:posttrain}), and the
three-view analysis uses a distinct 5{,}177-dialogue lineage whose genealogy is
in the supplement.

\paragraph{Audits: leakage and grounding.}
\label{sec:leakage}
Repair and regeneration are the paths by which benchmark content can reach
training data, so we audited them. A non-exhaustive sweep of four known-risk
packages recovers paraphrase clusters of 12--19 dialogues, each traceable to a
development error; in one condition 59 of 3{,}360 source dialogues recur as 118
of 4{,}517 exposures, and a three-seed removal rerun was too unstable to identify
an overlap effect. The separate 5{,}177-dialogue lineage contributes 112
dialogues sharing an eight-word span with five development cases. These are lexical screens
rather than semantic classifiers, and they demonstrably missed targeted
paraphrases, so they establish neither isolation nor decontamination. Grounding
was audited separately: roughly 19\% of earlier \code{Reason} fields asserted
findings the caller never supplied, which a verdict-preserving pass corrected,
and a later sweep removed 222 dialogues and regenerated 219 label-matched
replacements for them, leaving three removed outright.

\subsection{Construction Strategies and Their Evidence}
\label{sec:strategies}

A rule table admits many dialogue corpora, and which one gets built is a design
decision at every step. We describe what each strategy was meant to do and what
the records can say about it. None of the eight is separately identified: two
specify the labeling mechanism rather than add a testable component, two return
nulls, three are confounded with a co-occurring change, and one is evaluated
only at package level. The space itself, with its nulls and confounds, is the
finding here---not a per-strategy effect estimate.

\paragraph{Compiled labels, boundary pairs, and repair.}
The three strategies of \cref{sec:compile} carry different evidential weight.
Cited-row assignment is the labeling mechanism itself, so it admits no ablation.
One-fact pairs target tier boundaries, yet removing or replacing the 160 tagged
dialogues moves agreement by at most 2 of 201 cases, which establishes no
separate effect. Metadata-only repair remains a specification rather than a
verified property, because historical batches violated it.

\paragraph{Coverage without prior drift.}
A 309-sample phrasing increment rotated six styles across the nine symptom
categories that increment's own log recorded---a finer partition than the eight
generation families---without populating every category--style cell. On a
separate 191-case development set, accuracy rose from 69.8\% to 71.2\% and
emergent recall from 59.0\% to 83.0\%, while routine recall fell and over-triage
rose. The increment also added 136 emergent examples, so this was never a
style-only intervention, and a later one-for-one replacement held corpus size
fixed but yielded no isolated accuracy gain.

\paragraph{Evidence gating and prompt alignment.}
One configuration gated family-specific control behind posterior support and, in
the same change, aligned training with the deployment prompt. Two
question-quality scores logged by that harness both rose (51.9\%$\to$63.5\% and
35.9\%$\to$53.2\%), but the rubric behind them was not retained, so neither
quantity is defined here and we read only their direction. The two changes were
likewise never separated.

\paragraph{Label repair and negative results.}
Label repair neutralized 491 dialogues whose stated tier contradicted their
symptom family, corrected 91 carrying three mutually inconsistent labels, and
removed dialogues in which the target tier was recoverable from the prompt
rather than the caller's answers; on its filtered clean subset, epoch~2 reached
71.8\% agreement and 76.0\% emergent recall (\cref{sec:analysis}). The remaining
strategies show how easily a corpus change trades one property for another.
Aggressive pruning shifted the tier prior. 
A threshold change shipped together with a routing fix, so no separate effect is identified for either; Figure 5 \cref{sec:analysis} plots them as successive variants because that is the order the corpora were built, not because the changes were isolated.
Case-bank augmentation---adding 236 guideline-derived cases---raised early
emergent recall to 82.6\% while lowering accuracy to 63.7\% and pushing
over-triage to 27.4\%. 

\section{Experimental Setup}
\label{sec:eval}

\subsection{Benchmark and Reference Labels}

Human judgment is absent from training and concentrated in evaluation. The
internal benchmark converts 201 public, patient-authored ophthalmic Reddit
narratives into interactive profiles, omitting handles and URLs. The narratives
and the final reference were fixed before any comparison was run, but the same
cases had informed rule-table revision, corpus development, and model selection.

Labels were reviewed by exception rather than case by case. The workflow flagged
25 original labels and retained the other 176 by default, making the outcome a
candidate checklist rather than 201 independent decisions. Two rule-conditioned
LLM audits, prompted without the original labels, converged on the same
alternative for eight cases, and author adjudication revised 12, leaving 42
emergent, 47 urgent, and 112 routine references. We call the result the
\emph{operational reference} to keep its provenance visible: it was adjudicated
by the authors, not by an independent panel of clinicians, and its dependencies
run one way---the benchmark informed \Phirule{}, which informed both adjudication
and training. Internal agreement therefore tests the construction recipe for
policy conformance, not for clinical validity.

\subsection{Scoring the Three Views}
\label{sec:eval-decomp}

Because the questions a system asks determine which evidence exists at all, a
single case supports three readings against the same reference: the disposition
$v(h_i)$ the system states, \Phihat$(h_i)$ on the transcript it elicited, and
\Phihat$(x_i)$ on the complete narrative. We score all three on a separate
dialogue-revised condition---an earlier corpus revision, also $n{=}201$ but
distinct from Table~\ref{tab:headline}, 
on which GAO-Triage matches the reference on 147 of 201 cases---chosen because it is free of the defect audited below. The two rule views already disagree on 52
cases. Agreement on urgent-reference cases is higher from elicited transcripts
than from complete narratives (42.6\% versus 38.3\%), since the extra fields
extraction recovers can trigger emergent rules. The 72
matcher--reference disagreements separate by cause: the model is right on 29 of
33 extraction faults but on only 19 of 39 compiled-policy shortfalls. Since the
reference is policy-coupled, this audits a lossy matcher under two information
views and estimates neither an SFT effect nor generalization.

\subsection{Historical Protocols and Their Defects}

Several historical base--SFT comparisons carry a defect serious enough to
disqualify them as evidence. Retained code shows that the complete-profile
heuristic read the caller opening together with the hidden profile and, whenever
its own tier disagreed with the model's verdict, silently overwrote both the
final nurse turn and the stored disposition. In one archived historical protocol
(Protocol~B; the others are described in the supplement) we detect 31 base and 18
SFT overwrites, and an exact-template audit spanning 5{,}427 case-runs finds 19
to 44 overwritten cases per run. A milder issue affects the same runs: they share
the simulator's identifier and temperature yet sometimes ran days apart with no
immutable provider snapshot. Since no pre-overwrite verdict survives, the
metadata supports only bounds on direct agreement, nothing sharper.

\paragraph{Bounding pre-overwrite agreement.}
What the records still permit is a bound. The heuristic only ever fired on
disagreement, so an overwritten case that now agrees must previously have been
wrong, whereas an overwritten disagreement may have been right. Writing $C$ for
stored agreement and $M_+$, $M_-$ for the overwritten agreements and
disagreements, direct agreement lies in $[C-M_+,\;C-M_++M_-]$, assuming
overwriting was the only record change, that detection is exact, and that cases
factorize. Binary labels would identify the direct confusion matrix outright;
three tiers generally do not, and detection cannot be verified from the very
records it overwrote, so relaxing that assumption widens the range (general
bound and proofs in the supplement). One further asymmetry limits these runs:
training used symptom-family-specific prompts, while the retained evaluation
prompts were generic and unversioned.

\paragraph{Metrics and uncertainty.}
Against the operational reference we report exact agreement, reference-class
recall, and confusion matrices, with Cohen's $\kap$ \citep{cohen1960} and
balanced accuracy in the supplement. Every triage error we count, severe
under-triage
(\E{}$\to$\R{}, \U{}$\to$\R{}) included, is policy-relative rather than a
clinical outcome. Only one comparison is primary---base against SFT---with
endpoints and decision rules fixed in a written plan before scoring; that plan
carries no external timestamp, making it pre-specified rather than registered.
Every other $p$-value, in Table~\ref{tab:headline} and the supplementary
one-factor checks alike, is secondary, uncorrected, and descriptive. The design
is also underpowered for small effects: at $n{=}201$ and the observed discordance
rates, paired differences below roughly 10 points go undetected at 80\% power.
We repeated the primary contrast once with a second seed and once with a second
patient simulator, not in a crossed design, and every other contrast rests on a
single seed and simulator. Supplementary three-seed intervals resample seed and
case jointly and are descriptive, not confidence intervals over a seed
population.


\section{Results}
\label{sec:results}

\begin{table}[tb]
\centering
{\small
\setlength{\tabcolsep}{0.8pt}
\begin{tabular*}{\columnwidth}{@{\extracolsep{\fill}}lrrrr@{}}
\toprule
Nurse & E rec. & No v. & Agr. & $p$ \\
 & \% (hit/42) & $n$ & \% & vs.\,ours \\
\midrule
\textbf{GAO-Triage (ours)} & \textbf{69.0 (29)} & 1 & \textbf{74.1} & --- \\
\quad$\hookrightarrow$ rerun, seed 2 & 78.6 (33) & 0 & 73.1 & $0.878$ \\
Base 9B & 9.5 (4) & 0 & 61.7 & $0.0046$ \\
ROUTINE-only (floor) & 0.0 (0) & 0 & 55.7 & $3.8$e$\text{-}5$ \\
\midrule
\multicolumn{5}{@{}l}{\emph{Post-training: outcome GRPO ($\dagger$)}} \\
Outcome GRPO & 42.9 (18) & \textbf{80} & 45.3 & $1.9$e$\text{-}9$ \\
\midrule
GPT-5.5 & 64.3 (27) & 0 & 75.1 & $0.899$ \\
\quad$\hookrightarrow$ given the rule table & 66.7 (28) & 0 & 67.7 & $0.130$ \\
GPT-5.4 & 64.3 (27) & 0 & 70.6 & $0.464$ \\
DeepSeek-V4-Pro & 40.5 (17) & 0 & 71.6 & $0.590$ \\
GLM-5.2 & 45.2 (19) & 0 & 69.2 & $0.295$ \\
Qwen3.5-397B & 73.8 (31) & 0 & 68.2 & $0.141$ \\
Kimi-K2.6 & 76.2 (32) & 2 & 71.1 & $0.532$ \\
\bigqwen{} & 35.7 (15) & 0 & 62.2 & $0.0043$ \\
\bottomrule
\end{tabular*}
}

\caption{Clean results on 201 cases (42 emergent). Rows marked
$\hookrightarrow$ are variants of the row above them, not separate systems:
\emph{rerun, seed 2} retrains GAO-Triage under an identical recipe with a
different random seed, and \emph{given the rule table} supplies GPT-5.5 with the
same 70 rules our corpus was compiled from. E rec.\ is emergent recall (\%,
hit/42); No v.\ counts dialogues ending without a disposition; Agr.\ is exact
agreement; $p$ compares each row against GAO-Triage. No external arm dominates
GAO-Triage on (Agr., E rec.); GPT-5.5's 75.1\% versus 74.1\% is null
($p{=}0.899$), not equivalent. ROUTINE-only always answers routine---a
majority-class floor, not a system. Per-tier confusion breakdowns, including
catastrophic emergent-to-routine misses, are in the supplement. ($\dagger$) GRPO
descends from original SFT weights that were not retained, so it is not comparable to the
seed-matched run above.}
\label{tab:headline}
\end{table}

\subsection{The Primary Comparison}
\label{sec:clean-rerun}

The primary estimate pairs SFT against the untuned backbone in a same-day clean
evaluation. The original SFT weights were not retained, so we retrained the
3{,}000-dialogue condition from its archived configuration and SHA-256--verified
corpus under one seed, disabling the complete-profile heuristic for both arms.
Agreement rises from 61.7\% to 74.1\% (exact McNemar
$p{=}0.0046$; paired difference 12.4 points; asymptotic paired risk-difference
95\% CI 4.3--20.6 percentage points). The gain is concentrated rather than diffuse: emergent recall
carries it, moving from 9.5\% to 69.0\% (26 versus 1 discordant; unadjusted
$p{=}4.2{\times}10^{-7}$, which remains significant under Holm correction across
the three class-level tests), while urgent and routine differences cancel. Severe
under-triage falls from 24 to 17 cases, a descriptive count we did not test, and
per-tier confusion breakdowns are in the supplement.

Two checks bound how far this estimate depends on the particular run that
produced it. A second same-recipe seed reproduces the effect (73.1\%), and
repeating the pair with a second vendor's caller model (DeepSeek-V4-Pro,
temperature 0), everything else fixed, again preserves the gap through emergent
recall (71.4\% versus 7.1\%).

\paragraph{General-purpose models as nurses.}
\label{sec:external-reference}
Table~\ref{tab:headline} and Figure~\ref{fig:results} place GAO-Triage beside
eight external arms spanning seven general-purpose models, all under the same
nurse role, reference, and patient protocol. Only zero-shot \bigqwen{} agrees on
significantly fewer cases ($p{=}0.0043$); a 9B model trained on compiled guidance
is otherwise not separated from frontier systems here. Two confounds bound that
reading: the arms differ in interaction length (means 3.8--20.9 messages), and
GPT-5.5 shares a model family with the patient simulator. The ordering is also
reference-specific---it reverses on an external set, though not robustly
(\cref{sec:analysis}).

\subsection{Ablating the Construction Strategies}
\label{sec:ablations}

These strategy-level checks are underpowered, and we report them as such. The
boundary-pair null of \cref{sec:strategies} neither proves the strategy useless
nor detects a separate effect: at $n{=}201$ with one seed per arm, any effect
below the roughly ten-point threshold of \cref{sec:eval} would go unresolved. Supplying the rule table to
GPT-5.5 likewise yields no improvement---the point estimate in fact falls, 67.7\%
against 75.1\%---and repair has no on/off arm. 
Exact one-factor results and
comparators appear in Supplementary Section~C.

\subsection{Post-Training with Outcome GRPO}
\label{sec:posttrain}

Rewarding the disposition directly is the obvious next step, and it fails in an
instructive way. The GRPO checkpoint adds no human dialogue labels and was
re-evaluated clean, but it descends from original SFT weights that were not retained, whereas
Table~\ref{tab:headline} reports a same-recipe retrain, so the contrast compares
checkpoints rather than interventions from a common initialization. Agreement
falls to 45.3\% ($p{=}1.9{\times}10^{-9}$), and the cause is not misrouting but
silence: 39.8\% of dialogues end with no disposition, against at most two cases
for every other arm. These are early terminations rather than dialogues that ran
to the turn cap without deciding, and we count them as failures. The reward also carried rule, format,
and judged-process components, so the collapse cannot be pinned on the outcome
term alone. Heuristic-enabled logs had masked this failure entirely.

\section{What Carries the Signal}\label{sec:analysis}

\begin{figure*}[t]
\centering
\includegraphics[width=\textwidth]{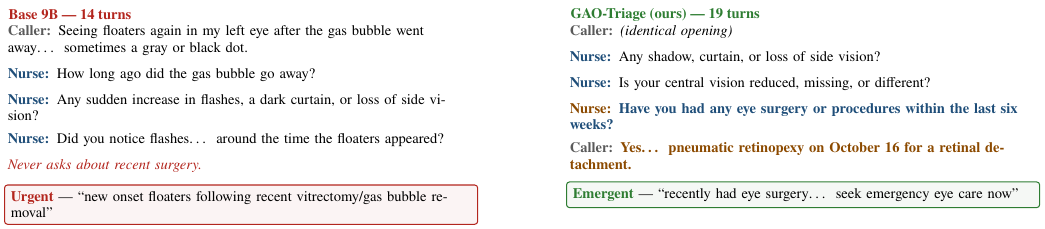}
\caption{One paired case under the clean protocol (reference tier \E{}); excerpts
are verbatim, shortened by ellipses. The trained arm screens for recent
intraocular surgery and elicits the decisive fact; the untuned arm never asks,
yet cites surgery in its own reason. The gain is in what gets asked. Bold marks the decisive exchange.}
\label{fig:case}
\end{figure*}

\begin{figure}[htbp]
\centering
\includegraphics[width=\columnwidth]{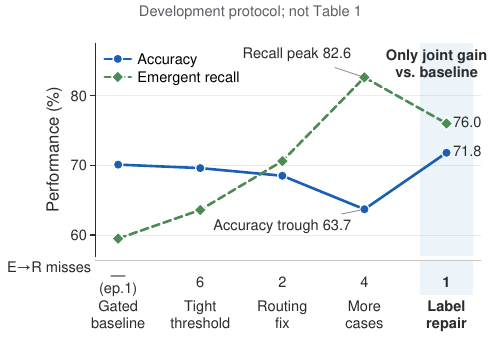}
\caption{
Late-checkpoint diagnostics across five successive development variants; E$\rightarrow$R totals span three epochs, except the gated baseline, for which only an epoch-1 count survives. Only label repair lowers the miss total and raises agreement together. These are successive variants on unequal filtered sets, not controlled replications.}
\label{fig:denoise}
\end{figure}

\paragraph{Where the signal lives.}
A corpus built this way could teach two things: the surface form of a triage
dialogue, or the guideline's mapping from evidence to tier. Shuffling tells them
apart. Permuting only which dialogue receives which label, holding class
marginals fixed, collapses the model onto \R{} for every case---agreement
$112/201{=}55.7\%$, recall $0/42$ for \E{} and $0/47$ for \U{}, exactly the
majority-class rate. Format, length, and tier frequency are therefore
insufficient alone. The control is not strict, since it used a 2{,}635-sample
corpus rather than the headline arm's 3{,}998 weighted exposures.

\paragraph{What the gain looks like case by case.}
Figure~\ref{fig:case} shows the same mechanism on one paired \E{} case: what SFT
buys is not better guessing from the same evidence but the acquisition of the
decisive evidence. One transcript illustrates the mechanism without establishing
it for the $4/42$ versus $29/42$ gap.

\paragraph{Safety is not read off agreement.}
The two axes of Figure~\ref{fig:results} come apart at the checkpoint level too.
On the same clean protocol, catastrophic \E{}$\rightarrow$\R{} counts vary widely
among arms of comparable agreement, and between two same-recipe seeds. Even
within one lineage the most accurate checkpoint is not the safest: case-bank
augmentation (\cref{sec:strategies}; ``More cases'' in
Figure~\ref{fig:denoise}) posts epoch-wise counts of $0/23$, $3/24$, and $1/27$
on separately filtered subsets, and its 72.7\% peak accuracy falls on the middle
epoch---the one that misses most. Total agreement therefore does not order
systems
by safety, and the aggregate base--SFT difference is pinned down far better than
any checkpoint's operating point.

\paragraph{What the development record cannot show.}
The variants in Figure~\ref{fig:denoise} are historical development records, not same-denominator main-protocol results: the gated baseline used all 201 cases, whereas in each later variant the complete-profile heuristic had touched 19\% to 44\% of cases in at least one of the three epoch runs (per-run counts are lower; Section 4.3), and every touched case was dropped, leaving unequal $n{=}111$--161 subsets with 17–27 emergent cases.
Each step also bundles co-occurring changes—the routing fix shipped with a threshold change, and label repair moved several edits at once—so directions are more trustworthy than magnitudes.
Related diagnostics stay in the supplement.

\paragraph{Late-checkpoint misses and label repair.}
Along that line, every threshold and coverage change we tried left the
catastrophic-miss count elevated, and case-bank augmentation bought emergent
recall at the price of accuracy and over-triage. Only label repair, applied to
the otherwise fixed routing-fix recipe (\cref{sec:strategies}), lowered the miss
total and raised agreement at once. The evidence is package-level: the edits were
never separately ablated, and the two miss counts rest on different denominators
(2/64 versus 1/71).

\paragraph{External stress tests.}
Every result so far is scored against a reference our own policy informed, so we
also tested against labels we did not write. On Oxford-40, an externally authored
expert-reviewed vignette set \citep{mittal2026ophthalmic}, a trained precursor of
our lineage---not the Table~\ref{tab:headline} condition---and the untuned
backbone agree with its 40 labels on 62.5\% and 50.0\%, while zero-shot GPT-5.5
reaches 82.5\% and reverses the within-reference ordering. The contrast carries
little weight either way: neither difference is robust 
($p{=}0.383$, trained versus base; $p{=}0.077$ Holm-adjusted, GPT-5.5 versus trained), the prompts differ, a constant-emergent baseline already attains
57.5\%, and Oxford-40 scores static vignettes rather than elicitation.

\paragraph{Limitations and ethics.}
The reference is the weakest link in every number above. Two rule-conditioned
LLM reviewers agreed on only 44.1\% of 930 citation-to-tier recomputations
($\kap{=}0.285$), and nine of twelve label revisions were downgrades, so it may
systematically under-call emergent cases. Audited packages also contain
paraphrase clusters, verdict parsing is permissive, and most arms rest on one
seed and one simulator. 
Frozen multi-clinician labels and an interactive external benchmark are prerequisites for any patient-facing use.
The data raise a separate concern. Training
dialogues are synthetic, but an unrecorded fraction of profile seeds and the
benchmark derive from public, patient-authored Reddit posts, and public posting
is not consent to machine-learning use. We retain no platform agreement,
acquisition record, or de-identification audit, leaving re-identification risk
unassessed; post-derived text reached remote endpoints whose retention policies
we did not record, and no IRB review was sought. The artifact withholds case
text; the system must not guide care.

\section{Conclusion}
\label{sec:conclusion}

We show that a compiled specialty guideline can stand in for instance-level dialogue annotation when training a multi-turn triage agent. A 70-row guideline table labels 3,000 dialogues without human judgments, improving operational-reference agreement from 61.7\% to 74.1\% and emergent recall from 9.5\% to 69.0\%. These gains require no frontier model at inference time, and none of the seven general-purpose systems we test dominates GAO-Triage on both metrics. Across eight guideline-to-dialogue strategies, two constitute the labeling mechanism itself, two yield null results, three remain confounded, and one is evaluated only as a package. Shuffling label–dialogue assignments collapses performance to the majority class, indicating that the capability comes from guideline-derived assignment rather than dialogue realism. Guideline compilation therefore offers a practical alternative to per-instance annotation where published guidance is available but annotated conversations are scarce. These results establish policy conformance, not clinical accuracy; independent multi-clinician labels and an interactive external benchmark are what would turn them into a clinical claim.


\bibliography{refs}

\end{document}